\documentclass[letterpaper]{article}
\usepackage{natbib,alifeconf}
\usepackage{graphicx}

\title{The Case for a Mixed-Initiative Collaborative Neuroevolution Approach}
\author{Sebastian Risi, Jinhong Zhang, Rasmus Taarnby, Peter Greve, Jan Piskur, Antonios Liapis, Julian Togelius\\
\mbox{}\\
IT University Copenhagen, 2300 Copenhagen, Denmark \\
$\{$sebr, jinh, reta, pgre, japi, anli, juto$\}$@itu.dk}

\begin{document}
\maketitle

\begin{abstract}
It is clear that the current attempts at using algorithms to create
artificial neural networks have had mixed success at best when it
comes to creating large networks and/or complex behavior. This should
not be unexpected, as creating an artificial brain is essentially a
design problem. Human design ingenuity still surpasses computational
design for most tasks in most domains, including architecture, game
design, and authoring literary fiction. This leads us to ask which the
best way is to combine human and machine design capacities when it
comes to designing artificial brains. Both of them have their
strengths and weaknesses; for example, humans are much too slow to
manually specify  thousands of neurons, let alone the billions of
neurons that go into a human brain, but on the other hand they can
rely on a vast repository of common-sense understanding and design
heuristics that can help them perform a much better guided search in
design space than an algorithm. Therefore, in this paper we argue for a mixed-initiative approach for collaborative online brain building and present first results towards this goal.  

\end{abstract}

With around 200 billion neurons and 125 trillion synapses, the human brain is the most complex system known to exist. The brain's structural complexity, with intricate synaptic motifs that repeat throughout it, gives rise to our unique mental abilities. Additionally, the brain’s plasticity allows us to learn new abilities throughout our life (e.g.\ learning a new language) and to change our behavior based on past experience.

Therefore, creating similar artificial structures by recapitulating the process that created intelligence on earth, is an intriguing possibility. In this context, neuroevolution (i.e.\ evolving artificial neural networks (ANNs) via evolutionary algorithms) has shown promising results in a variety of different domains \citep{floreano08neuroevolution,stanley:jair04,yao_evolving_1999}. However, these results still pale in comparison to the capabilities of natural brains. The reasons for this are manifold. Especially the problem of deceptive fitness landscapes (i.e.\ mutations increase fitness but actually lead further away from the final objective) has limited the scope of problems amenable to evolutionary algorithms. 

Research in circumventing this problem has mainly focused on two different ideas. First, novelty search \citep{lehman:ec11}, a method that rewards novel behaviors instead of rewarding objective performance, has shown promise and significantly outperforms objective-based approaches in a variety of different domains. Other approaches are based on interactive evolutionary computation (IEC) methods, wherein the human user guides evolution by repeatedly choosing from a set of candidates. \citet{woolley2011deleterious} showed that interactive evolution can help to discover artifacts which are very hard to evolve with traditional evolutionary approaches. Recently \citet{woolley2014} combined IEC with novelty search, demonstrating that the approaches complement each other and together address some of the challenges that each method struggles with by itself (e.g.\ novelty search can get lost in large search spaces, interactive evolution is limited by user fatigue). In a related approach, \cite{bongard2013combining} recently demonstrated that human input and objective-based search can also be combined synergistically to solve challenging robotic domains.

While these \emph{human in the loop} approaches have shown promising results,  we argue in this paper that they do not exploit the whole range of human intuition, ability to identify promising stepping stones, or collaborative problem-solving skills that could ultimately allow us to create more brain-like artificial structures. For example, collaborative games like Foldit 
hint at the power of crowdsourcing the human brain's natural abilities for specific tasks (e.g.\ pattern matching, spatial reasoning), which are hard to solve by computational approaches. However, in traditional IEC applications the role of the user is often reduced to solely judging the created artifacts and only ``nudging'' evolution by deciding between a discrete choice of candidates. In other words, only the computer creates content (e.g.\ images, ANNs, etc.) and the role of the human is to guide evolution to content they prefer. 

An approach that does require significantly more input from the user is a \emph{mixed-initiative process} \citep{Liapis2014mia}, wherein both the computer and the human take turns in creating content. Yet while mixed-initiative based approaches have shown promise in the context of procedurally generated content for games \citep{liapis2013sentient,yannakakis2014micc}, they have not yet been applied to the evolution of large-scale ANNs for complex tasks. However, the promise of such an approach is a system that can benefit from the different skillsets of a human and a computational creator. For example, while a human user might develop an intuition about promising domain-dependent network topographies, a computational method is likely more effective at fine-tuning specific synaptic weights. Therefore, we argue for a collaborative mixed-initiative approach in which both the computer and the collaborating human users can take initiative and propose changes to an evolving neural networks. That is, at any point the human can revert to just doing selection and let the evolutionary algorithm serve up new content to judge, or decide to jump in and have a more active role in the design of the ANN. 

\begin{figure}
\centering
\includegraphics[width=3.3in]{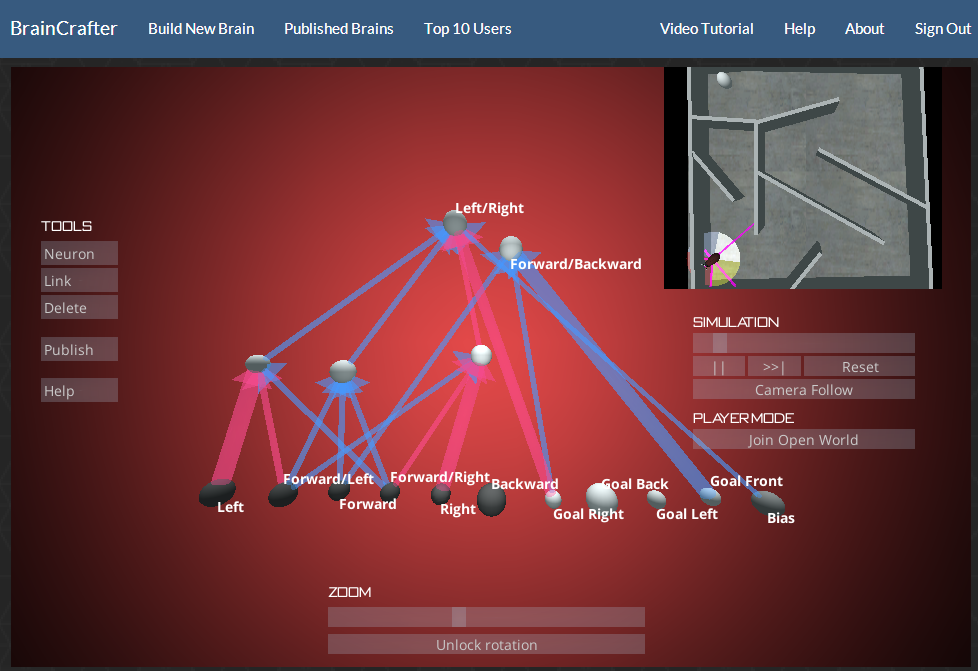}
\caption[]{BrainCrafter Web Interface. Beta version online at: braincrafter.dk}
\label{fig:brain_builder}
\end{figure}

Our recent work has taken steps towards this goal by focusing on two parallel lines of research. First, before introducing a mixed-initiative approach it is useful to determine how good we as humans are at building complex ANNs for certain tasks (without evolution) and if collaborating with other users proves useful. Insights from this experiment should in turn provide useful clues about the strengths of human ANN design and most importantly, non-intuitive aspects of the design process we tend to struggle with (i.e.\ aspects which would benefit most from an assisting computational creator). In this context we recently introduced \emph{BrainCrafter} (braincrafter.dk), which is an online application that allows the user to build ANNs for specific control programs (e.g.\ a robot that must traverse a maze) by adding neurons and connections in a drag$\&$drop like fashion (Figure~\ref{fig:brain_builder}). While building ANNs the users can observe the resulting simulated robot behaviors in real-time, proving insights into the effects of different network modifications. BrainCrafter also allows users to collaborate by building on high-scoring solutions created by other people. 

While the ongoing BrainCrafter experiment should provide insights into our abilities to collaboratively construct complex networks, more challenging domains will likely require the orchestrated effort of both human \emph{and} computational creators. Therefore extending BrainCrafter to support the users' collaborative engineering efforts through a mixed-initiative approach is an important next goal. To allow seamlessly switching between interactive evolution and manual ANN design will require an ANN representation that (1) can be edited easily by the user on a local (e.g.\ individual neurons) and global level (e.g.\ overall network topography and topology), (2) is evolvable and compact, and (3) allows the system to produce meaningful suggestions based on the user's input.

\begin{figure}
\centering
\includegraphics[width=3.3in]{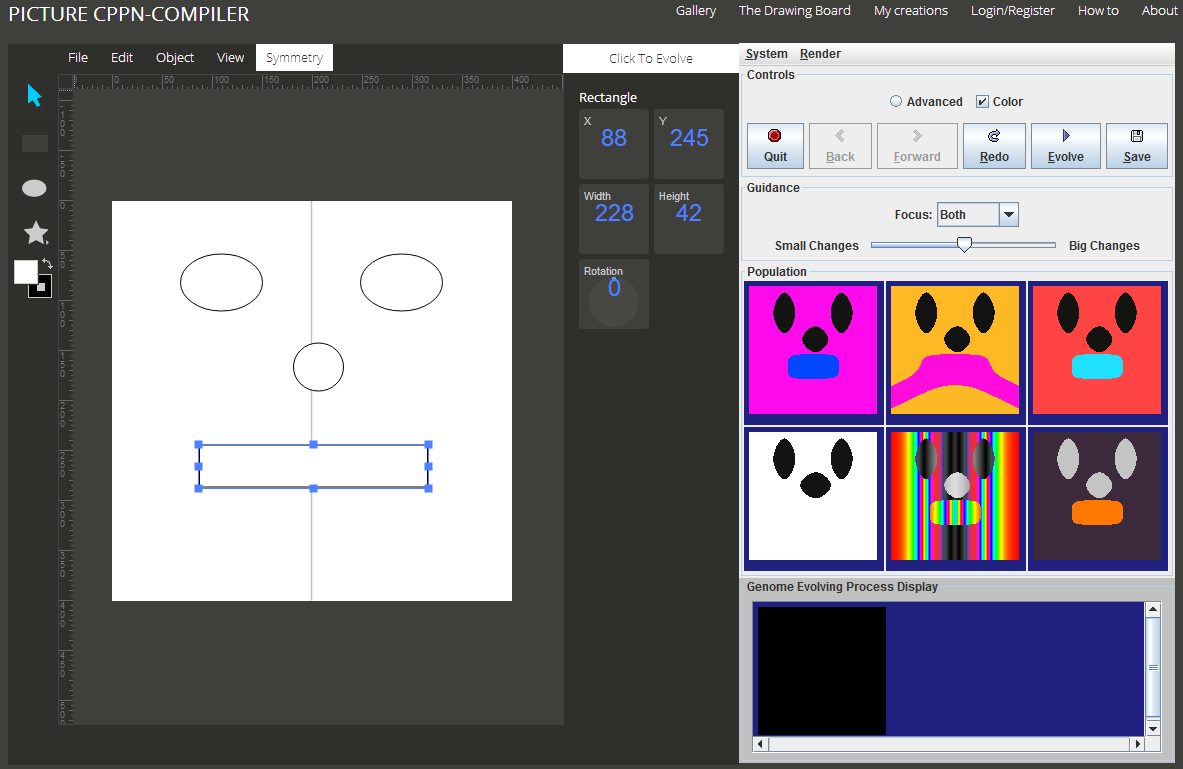}
\caption[]{Picture CPPN-Compiler User Interface. Beta version online at: rasmustaarnby.dk/thesis}
\label{fig:cppn_compiler}
\end{figure}

In this context we are building on the a generative encoding called \emph{compositional pattern producing networks} (CPPNs; \cite{stanley:gpem07}) and the recently introduced concept of a CPPN-Compiler \citep{risi2013compiler}. CPPNs, which are based on principles of how natural organisms develop, allow the compact encoding of complex patterns, from two-dimensional images \citep{secretan:ec11} and three-dimensional forms~\citep{clune:ecal11} to large-scale ANNs~\citep{stanley:alife09,clune2011performance,risi2012enhanced}. The idea behind the CPPN-Compiler is to allow the user to directly compile a high-level description of the desired starting structure into the CPPN itself. We are currently exploring the benefits of this approach through the collaborative interactive evolution of images (Figure~\ref{fig:cppn_compiler}), in which the user can draw a vector image and annotate it with important regularities like symmetry. The CPPN-Compiler then compiles this high-level description into the CPPN itself. Since the compiled CPPN now directly embodies the annotated domain regularities (e.g.\ bilateral symmetric arms), the produced offspring show meaningful variations that nevertheless share common features. Considering the insight that CPPNs can also produce large-scale ANNs \citep{stanley:alife09} and can be modified to create certain neural topologies (ES-HyperNEAT; \cite{risi2012enhanced}), opens up the intriguing possibility of a neural CPPN-Compiler. Our current efforts focus on such a compiler that will form the backbone for our mixed-initiative BrainCrafter application. 

\begin{figure}
\centering
\includegraphics[width=3.3in]{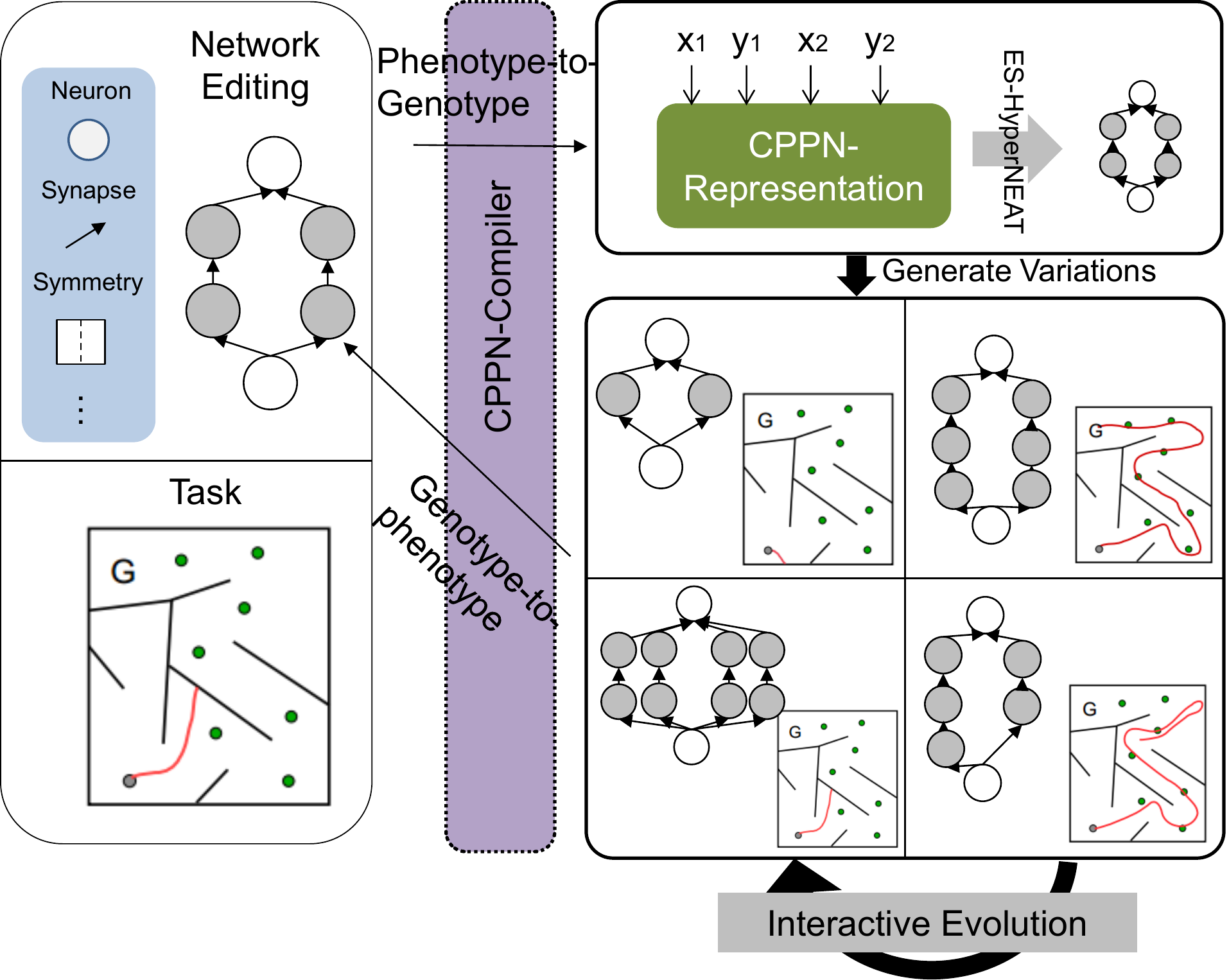}
\caption[]{Mixed-initiative Neuroevolution Framework}
\label{fig:overview}
\end{figure}

The envisioned collaborative mixed-initiative system is depicted in Figure~\ref{fig:overview}. The users can collaboratively construct neural networks and annotate them with regularities (e.g.\ symmetry, repetition, etc.), which are used by the computational creator to construct the internal CPPN model and in turn propose meaningful variations to the user. At any point in the process the human can revert to just doing selection or decide to directly edit the ANN, wherein phenotypic edits are directly compiled back into the genotype. The promise of the proposed system is that it could allow a variety of tasks to be solved by many people online within a mixed-initiative environment, which have heretofore proven too difficult. We expect this project will profoundly impact the fields of ANN research and potentially also deepen our understanding of the way biological neural networks solve certain problems.

\footnotesize
\bibliographystyle{apalike}
\bibliography{nnstrings,nn,ucf,multiagent,GECCO08,petalz,neuro}

\end{document}